# DISTRIBUTED EVOLUTIONARY COMPUTATION: A NEW TECHNIQUE FOR SOLVING LARGE NUMBER OF EQUATIONS

Moslema Jahan, M. M. A Hashem and Gazi Abdullah Shahriar

Department of Computer Science and Engineering,
Khulna University of Engineering and Technology, Khulna 9203, Bangladesh
mjahan.cse@gmail.com, mma_hashem@hotmail.com and rana_kuet@yahoo.com

### ABSTRACT

*Evolutionary computation techniques have mostly been used to solve various optimization and learning problems successfully. Evolutionary algorithm is more effective to gain optimal solution(s) to solve complex problems than traditional methods. In case of problems with large set of parameters, evolutionary computation technique incurs a huge computational burden for a single processing unit. Taking this limitation into account, this paper presents a new distributed evolutionary computation technique, which decomposes decision vectors into smaller components and achieves optimal solution in a short time. In this technique, a Jacobi-based Time Variant Adaptive (JBTVA) Hybrid Evolutionary Algorithm is distributed incorporating cluster computation. Moreover, two new selection methods named Best All Selection (BAS) and Twin Selection (TS) are introduced for selecting best fit solution vector. Experimental results show that optimal solution is achieved for different kinds of problems having huge parameters and a considerable speedup is obtained in proposed distributed system.*

### KEYWORDS

*Master-Slave Architecture, Linear Equations, Evolutionary Algorithms, Hybrid Algorithm, BAS selection method, TS selection method, Speedup.*

## 1. INTRODUCTION

In recent years, application of evolutionary algorithms is increasing to a greater extent. Evolutionary algorithms (EAs) are stochastic search methods that have been applied successfully in many search, optimization and machine learning problems. Successful use of evolutionary algorithm for solving linear equations is applied in [1], [2], [3]. However, it is often very difficult to estimate the optimal relaxation factor, which is the key parameter of the successive over relaxation (SOR) method. Optimal solution is achieved quickly as relaxation factors are adapted automatically in evolutionary algorithm. Equation solving abilities was extended in [2], [3], [4] by using time variant parameter. The invention of hybrid evolutionary algorithm [5], [6] brought a greater benefit to solve linear equations within very short time. Many problems with huge parameters such as Numerical Weather Forecasting, Chain Reaction, Astrophysics (Modelling of Black hole), Astronomical formation, Semiconductor Simulation, Sequencing of the human genome, Oceanography need high computational cost in case of single processor. One approach to overcome this kind of limitation is to formulate the problem into distributed computing structure.

The main parallel achievements in the algorithmic families including the evolutionary computation, parallel models and parallel implementations are discussed in [7]. A distributed cooperative coevolutionary algorithm is developed in [8] which is beneficial for solving complex problems. As there are no free lunch theorems for optimization algorithms, a graceful convergence is the key challenge for designing an optimization algorithm. A number of "no free lunch" (NFL) theorems [9] are presented for any algorithm, which state that any two algorithms are equivalent when their performance is averaged across all possible problems. On the other hand, there are coevolutionary free lunches theorems in [10]. The proposed technique follows coevolutionary theme. A distributed technique [11] is proposed for parallelizing fitness evaluation time. Fitness evaluation time is high but other operations of evolutionary algorithm take more time. This paper proposes a Distributed Evolutionary Computation (DEC) in which mutation and adaptation processes are also distributed. This is the Champion Selection technique, where best champion is selected within short period of time.

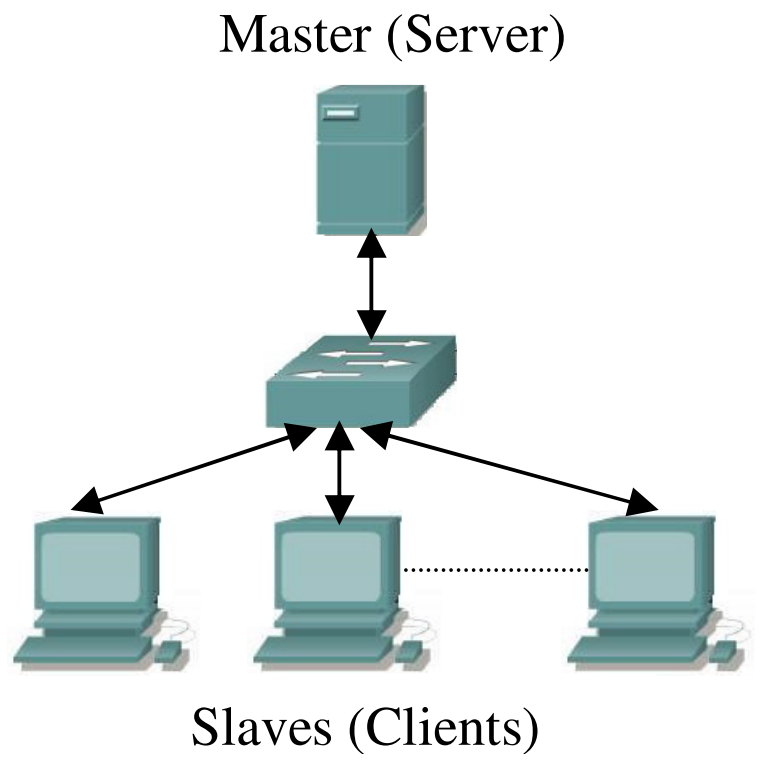

Figure 1. Distributed Master-Slave Architecture

Basic master-slave architecture (Figure 1) is used in proposed distributed technique that follows a server-client paradigm where connections are closed after each request. Slaves are connected with master through local area network (LAN) to take the advantage of distributed processing power of slaves. The basic approach of this system is to split a large problem into many subproblems and to evolve subproblems separately. These subproblems are then combined and actual solution is achieved. This process continues until the less erroneous solution comes out compared to a threshold error level.

The remainder of this paper is structured as follows: Section 2 represents the previous work related to the proposed work. Section 3 mentions Jacobi method of solving linear equations and distributed model. Timing calculation is discussed in section 4. Section 5 examines results of the experiment of various problems and provides a comprehensive comparison of single and distributed system on the basis of BAS and TS selection mechanism. Finally, Section 6 provides our concluding remarks.

## 2. RELATED WORK

The intrinsically parallel and distributed nature of EAs did not escape the attention of early researchers. Grefenstette [12] was one of the first in examining a number of issues pertaining to the parallel implementations of GAs in 1981. Grosso [13] is another early attempt to introduce parallelism using spatial multipopulation model. Several attempts were made to have a better and fast system that is capable of doing Evolutionary computations in parallel fashion. DREAM (Distributed Resource Evolutionary Algorithm Machine) [14] is such a system that used island model architecture on peer-to-peer connection. Both island-model and master-slave architecture has been combined at ParadisEO (PARAllel and DIStributed Evolving Objects) [15]. But in either case, all genetic operations are not done with a distributed manner. Among different models and architectures, this paper follows master-slave architecture to develop parallel and distributed environment.

JDEAL (Java Distributed Evolutionary Algorithms Library) [16] is a master-slave architecture coded in Java platform. Paladin-Dec [17] was another Java implementation of genetic algorithms, genetic programming and evolution strategies, with dynamic load balancing and fault tolerance. Still the communications among the nodes in distributed architecture uphold an issue. ECJ [18] is a Java-based framework that doing its computation using Java TCP/IP sockets. MPI (Message Passing Interface) is used at [19] with C++ framework. The developed distributed EC system was integrated transparently with the C++ Open BEAGLE framework [20] in 2002. Parallel performance of MPI sorting algorithms is presented in [21]. By gathering all ideas this paper implement a hybrid algorithm combining the jacobi-based successive relaxation (SR) method with evolutionary computation techniques which follow Java-based framework with socket programming in distributed manner and related speedup is calculated. Same algorithm was implemented in single processing system [4] using C++ framework and the related speedup was calculated. Finally experimental result shows distributed system is more speedy than single system to solve the problems with huge parameters.

Selection mechanism of an Evolutionary Computation technique has been a key part which brings up a significant computational cost. In 1982 Hector Garcia-Molina provides an election algorithm in [22] for two categories failure environments by which a coordinator will be selected when failure occurs in distributed computing system. A sum-bottleneck path algorithm is developed in [23] that allows the efficient solution of many variants of the problem of optimally assigning the modules of a parallel program over the processors of a multiple-computer system under some constraints on the structure of the partitions. An individual selection method is provided in [24] to select efficient individuals and whole system follows a distributed technique. In 2010, an improved system of an abstraction, all-pairs that fits the needs of several applications in biometrics, bioinformatics, and data mining is implemented in [25] shows the effect of campus grid system which is more secured than single system because of following distributed manner. Different papers followed various election mechanisms for choosing right population. Like these, our paper also provides two new selection mechanisms BAS which choose all best individual with less error rate and TS which choose one individual between adjacent two and develop twin copy of the individual with low error.

## 3. PROPOSED MODEL OF DISTRIBUTED EVOLUTIONARY ALGORITHM

The new distributed technique is anticipated based on Jacobi Based Time- Variant Adaptive Hybrid Evolutionary algorithm [3] through cluster computing environment. The proposed algorithm initializes a relaxation factor in a given domain which is adapted with time and fitness of the solution vector.

### 3.1. Jacobi Method of Solving Linear Equations

Consider the following linear equations:

$$\mathrm{Ax} = \mathrm{b} \text{ or } \mathrm{Ax} - \mathrm{b} = 0 \tag{1}$$

Where $A \in \Re^n \times \Re^n$ and $x, b \in \Re^n$

Let, A is $\mathrm{n} \times \mathrm{n}$ matrix, where $a_{ii}$ is the diagonal elements, $a_{ij}$ is other elements of the A matrix and $b_i$ is elements of $b$ matrix. For solving linear equations Jacobi method is used [26].

Let, Linear equations is $\mathrm{Ax} = \mathrm{b}$ and $|\mathrm{A}| \neq 0$

Then $(D + U + L)x = b$ , Where $\mathrm{A} = (D + U + L)$

or, $Dx = b - (U + L)x$ or, $x = D^{-1}b - D^{-1}(U + L)x$

or, $x = H_j x + V_j$

Where $H_j = D^{-1}(-L - U)$ and $V_j = D^{-1}b$

The linear equation can be written as

$$\sum_{j=1}^{n} a_{ij} x_j = b_j , \ (i = 1,2,3,.....n) \tag{2}$$

In Jacobi method by using SR technique [27] is given by,

$$x_i^{(k+1)} = x_i^{(k)} + \frac{\omega}{a_{ii}} \left( b_i - \sum_{j=1}^{n} a_{ij} x_j^{(k)} \right), \qquad (i = 1,2,3,.....n) \tag{3}$$

In matrix form matrix-vector equation is

$$x^{(k+1)} = H_\omega x^{(k)} + V_\omega \tag{4}$$

Where $H_\omega$ called Jacobi iteration matrix, and $V_\omega$ are given successively by

$$H_\omega = D^{-1}\{(1-\omega)I - \omega(L+U)\} \tag{5}$$

And $$V_\omega = \omega D^{-1} b \tag{6}$$

If $\omega$ is set at a value between 0 and 1, the result is weighted average of corresponding previous result and sum of other (present or previous) result. It is typically employed to make a non-convergence system or to hasten convergence by dampening out oscillations. This approach is called successive under relaxation. For value of $\omega$ from 1 to 2, extra weight is placed. In this instance, there is an implicit assumption that the new value is moving in the correct direction towards the true solution but at a very slow rate. Thus, the added weight $\omega$ is intended to improve the estimate by pushing it closer to the truth. This type of modification, which is called over relaxation, is designed to accelerate the convergence of an already convergent system. This approach is called successive over relaxation (SOR). The combine approach, i.e. for value of $\omega$ from 0 to 2, is called successive relaxation or SR technique [26].

Iterative process is continued using equation (3) until the satisfactory result is achieved. Based on this method, different hybrid evolutionary algorithms are developed in [1] [2]. Parallel iterative solution method for large sparse linear equation systems is developed in [28].

## 3.2. Proposed Distributed Technique

The availability of powerful network computers represents a wealth of computing resources to solve problems with large computational effort. Proposed distributed technique uses master-slave architecture and classical numerical method with Evolutionary Algorithm for solving complex problems.

In this system, large problems are decomposed into smaller subproblems and mapped into the computers available in a distributed system. Communication process between master and slaves follow message passing paradigm that is identical with [29]. This technique introduces the high performance cluster computation, linking the slaves through LAN to exploit the advantage of distributed processing of the subproblems. Here each slave in a cluster always solves same weighted subproblems although machine configuration is different. In this cluster computing environment, master processor defines number of subproblems according to slave number in a cluster. The workflow diagram of proposed algorithm is portrayed as in Figure 2.

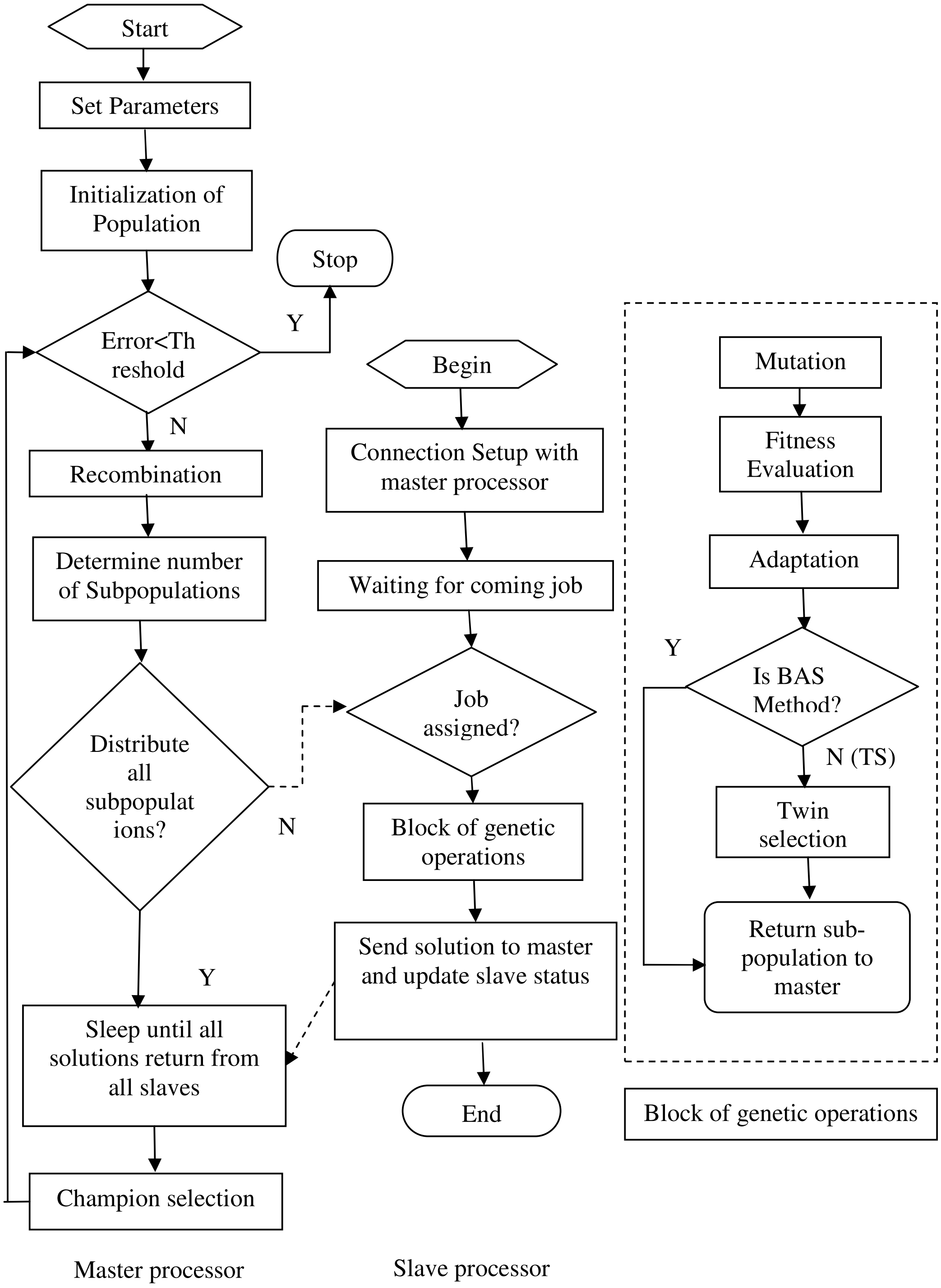

Figure 2. The workflow diagram of DEC system

The proposed workflow diagram can be mapped into master-slave paradigm. In Figure 3 all steps of proposed DEC are expressed using numbering system in each position. Step 1, 2, 7 are completed in master processor and step 3, 4, 5 occurred in slave processors but working principle of step 7 depends on selection method. For both methods, step 6 is dedicated for checking whether selected offspring from all slaves return back in master processor or not.

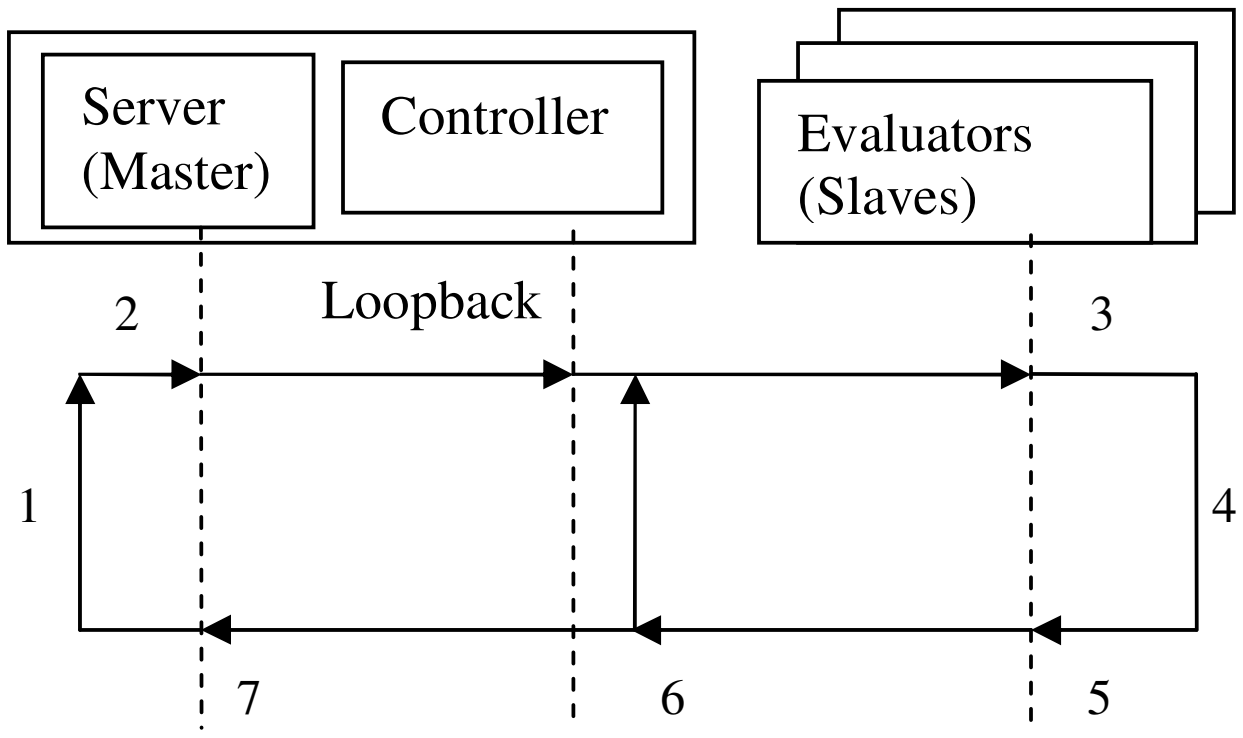

Figure 3. Working model of proposed technique

### 3.2.1. Initialization

In step 1, master processor initializes problems with population and relaxation factor. Initial population $X^{(0)} = \{x_1^{(0)}, x_2^{(0)}, .....x_N^{(0)}\}$ is generated randomly using normal distribution at master. Here $N$ is the population size. Let $k \leftarrow 0$ where $k$ is the generation counter. Relaxation factors $\omega^{(0)} = \{\omega_1^{(0)}, \omega_2^{(0)}, ......, \omega_N^{(0)}\}$ are also initialized on the basis of corresponding individuals.

### 3.2.2. Recombination

In step 2, Recombination operation is performed on parent at master and population $X^{(k+c)} = R(X^{(k)})^t$ is obtained, where $R = (r_{ij})_{N \times N}$ and $\sum_{j=1}^{n} r_{ij} = 1$ and $r_{ij} \geq 0$ for $1 \leq i \leq N$

Matrix $R$ is a stochastic matrix.

Population generated after recombination operation is distributed among slaves. Then mutation, fitness evaluation and adaptation operations are performed on that distributed subpopulations.

### 3.2.3. Mutation

After completing recombination operation step 3 provides Mutation operation on the subpopulation in slave processors and mutated subpopulation $X^{(k+m)}$ is obtained. For each Mutation on subpopulation is

$$\sum_{i=1}^{N_{sub}} x_i^{(k+m)} = H_\omega x_i^{(k+c)} + V_\omega \quad (7)$$

### 3.2.4. Fitness Evaluation

After completing mutation operation, fitness evaluation is performed in step 4. Error function is

$$E = \sum_{i=1}^{N_{sub}} E_i \quad (8)$$

$$\text{Where } E = \sum_{j=1}^{N_{sub}} \left| a_{ij}(x_j) - b_i \right|, \quad i = 1,2,....., N_{sub}$$

The fitness evaluation of an individual measures the accuracy of an individual for a particular problem and calculates the error rate which is used for selecting best individuals.

### 3.2.5. Adaptation

In step 5, adaptation is performed on mutated offspring according to fitness evaluation. Consider two individuals $X$ and $Y$, corresponding errors $\|e_x\|$ and $\|e_y\|$ and relaxation factors $\omega_x$ and $\omega_y$.

If $\|e_x\| > \|e_y\|$ then $\omega_x$ is adapted to $\omega_y$.

$$\omega_x^m = (0.5 + p_x)(\omega_x + \omega_y)$$

Similarly if $\|e_y\| < \|e_x\|$ then $\omega_y$ is adapted to $\omega_x$ and if $\|e_x\| = \|e_y\|$ then no adaptation is performed. $P_x$ is denoted as adaptive (TVA) probability parameter of $\omega_x$.

In step 6, controller checks whether offspring from all slaves have reached in master processor or not. In BAS method, step 6 starts its task after completion of adaptation operation but in TS method, after completing of partial selection of best offspring in each slave.

### 3.2.6. Selection

The DEC system provides two selection methods named BAS and TS. In BAS method, best individuals are selected among parent and mutated offspring. At position 7, mutated offspring from all slaves are combined and selection mechanism is performed on parent and newly generated offspring by mutation.

Mathematically,

$$Selection(i) = err_{\min}\{x_1, x_2, x_3, .....x_{2N-i+1}\} \quad (9)$$

$$Selection = \sum_{i=1}^{N} Selection(i) \quad (10)$$

Individuals compete among themselves and select best individual based on error value. Before finding the optimized solution, overall system will be continued in same process.

TS method provides a partial selection operation on mutated offspring in each slave where one best individual is selected between two consecutive individuals and developed twin copy of selected offspring.

Mathematically,

$$Selection(i) = err_{\min}\{x_i, x_{i+1}\} \quad (11)$$

$$Selection = \sum_{i=1}^{N_{sub}} Selection(i) \quad (12)$$

These selected offspring are combined in server and select best one. The selected copy will be the champion among all individuals or optimized solution for a particular problem if this fulfils the desired condition, otherwise fill up the whole archive and next generation will be continued.

There is no direct communication among the subpopulations. All communications are performed between the subpopulation and the central server. Such communications take place in every generation before reaching the accurate result.

## 3.3. Scenario of "Champion Selection"

The DEC system can be compared with a game where a champion will be selected on the basis of some criteria. At the starting moment, all players are presented at master which is a coordinator. The Coordinator divides all players into different teams and assigns these teams in different slaves. The number of players in a team is determined according to the slave number in a game region that is a cluster. After reaching the team, some operations are performed on that team separately and simultaneously in each slave. Each player has enriched after operation and all players from each slave come back in master. After coming back, all players compete with each other and select best one. If the selected one fulfils the desired criteria then this is the champion of the game or game will be continued following same process.

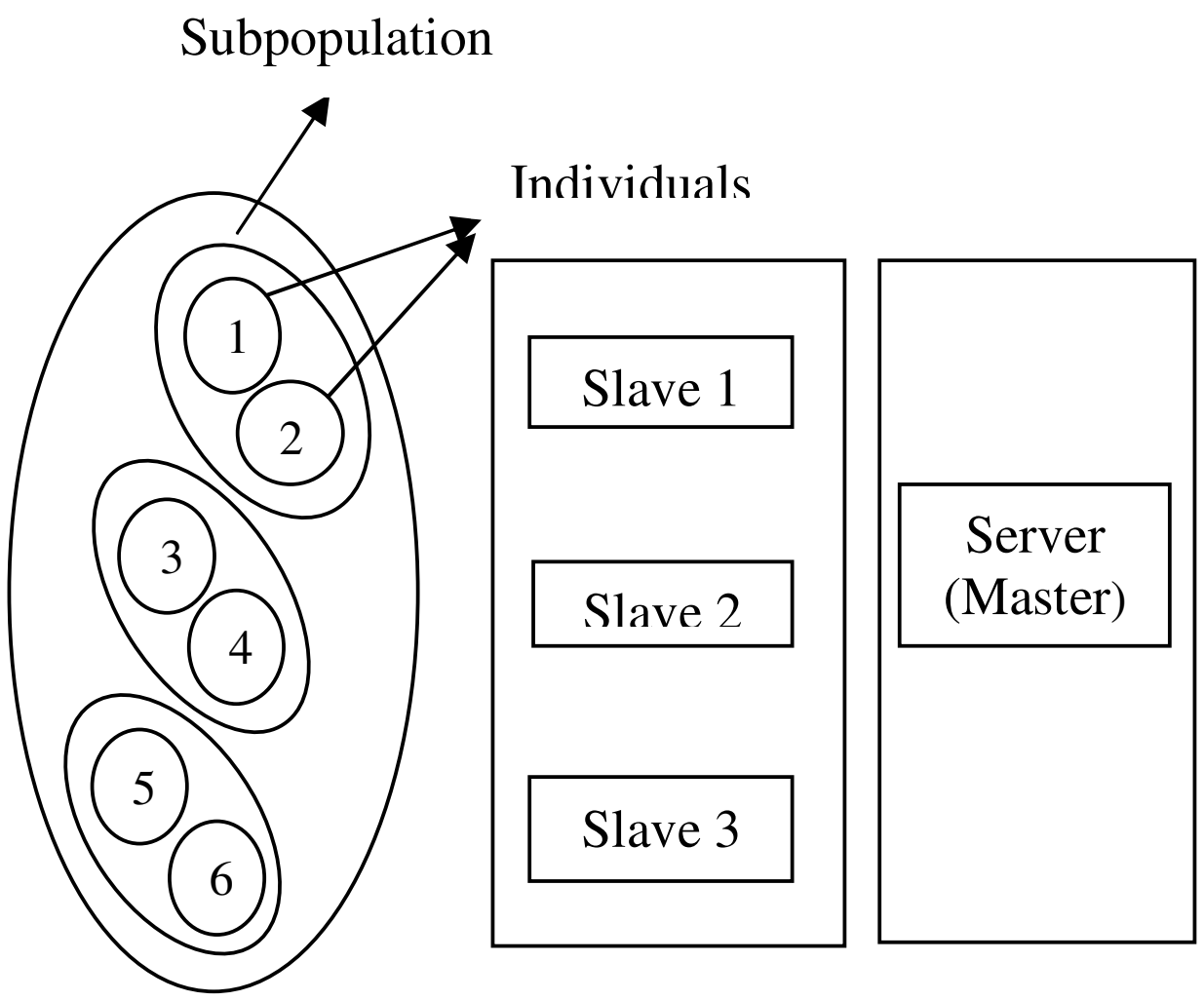

Figure 4. The model of DEC

A prototype model of DEC with six individuals over three slave computers is illustrated in Figure 4. The figure shows the division of main population into three subpopulations containing two individual in each according to the number of slave computers where master processor is the coordinator which will take the decision. Each slave accepts two individuals among six of them. Some Evolutionary mechanisms are operated separately in each slave. After completing the operations, subpopulations are returned back to master and individuals compete among themselves and a champion is coming out. If the champion is not best suited with the standard value, this process will be continued. Otherwise this champion is the winner of the game.

## 3.4. Explanation of Selection Methods

This section represents two selection methods which will help to find out the optimized solution of a problem. Working mechanisms of these two methods are as follows:

### 3.4.1. BAS selection method

In this method, best individual is selected between parent and mutated offspring in each iteration. After coming back all individuals from all slaves to master, they compete with each other and provide best half of the total individuals including parent and offspring and assign priority according to the error rate. On the basis of the priority, a champion is selected. If the error rate of the champion is equal or less to the standard value, this is the optimized solution of the problem. Otherwise the process will be going on.

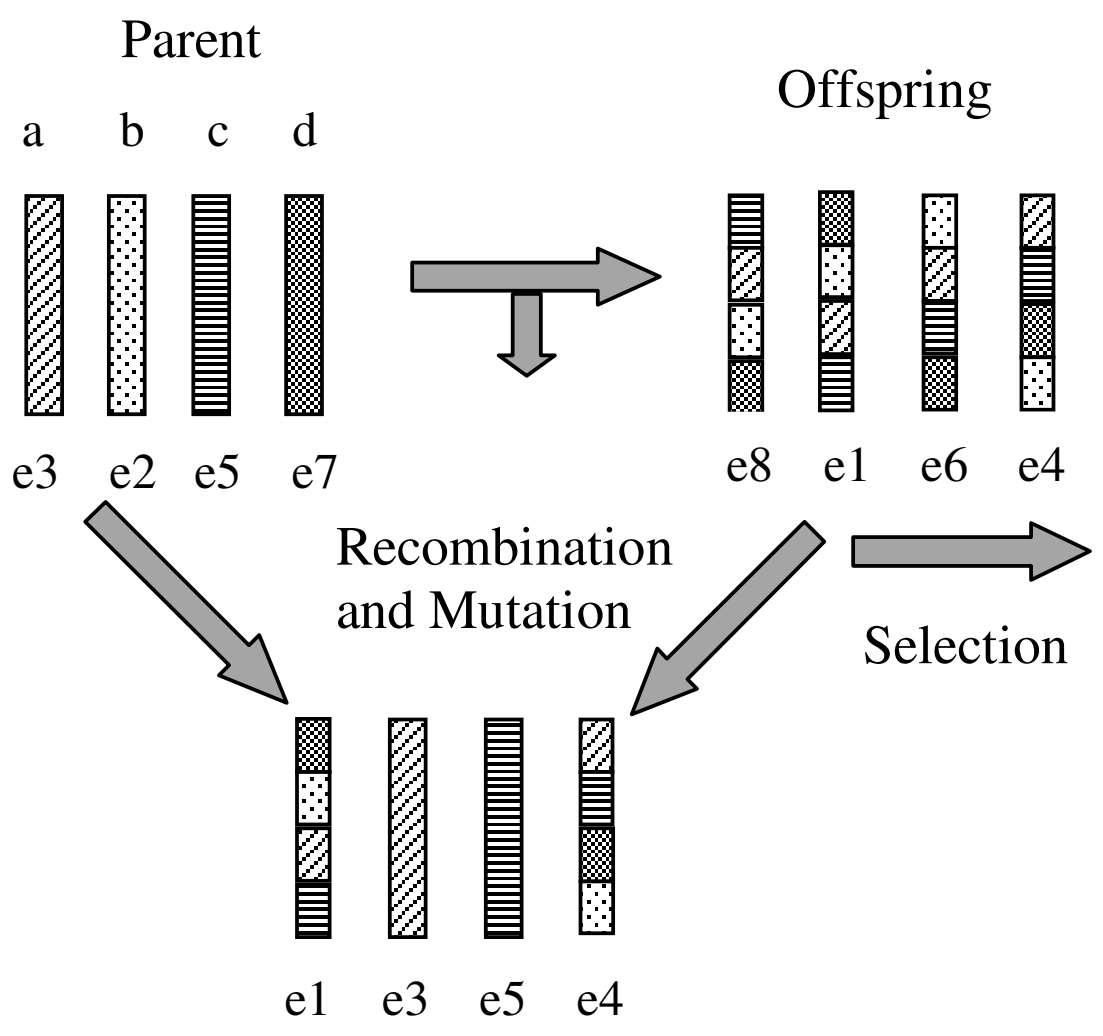

Figure 5. Selection mechanism for BAS method

In Figure 5: a, b, c and d patterns represent four parent individuals and the stripped patterns represent offspring. Error of each parent individual is e3, e2, e5, e7 and corresponding relaxation factors are ω3, ω2, ω5, ω7. Similarly error of each offspring is e8, e1, e6 and e4 and relaxation factors are ω8, ω1, ω6 and ω4. Here error number represents the value of the error. Now best half individuals are selected between parent and offspring according to error value. Here two individuals are selected from parent and two are selected from offspring. Corresponding relaxation factor is also rearranged. Next generation is started with these selected offspring and their respective relaxation factor.

The time variant adaptive (TVA) parameters are defined as

$$P_x = E_x \times N(0,0.25) \times T_\omega \tag{13}$$

And $P_x$ is denoted as adaptive (TVA) probability parameter of $\omega_x$, and

$$P_y = E_y \times [N(0,0.25)] \times T_\omega \tag{14}$$

And $P_y$ is denoted as adaptive (TVA) probability parameter of $\omega_y$; Where $T_\omega = \left(1 - \frac{t}{T}\right)^\gamma$

Here $\lambda$ and $\gamma$ are exogenous parameters, used for increased or decreased of rate of change of curvature with respect to number of iterations; $t$ and T denote number of generation and maximum number of generation respectively. Also $N(0,0.25)$ is the Gaussian distribution with mean 0 and standard deviation 0.25.

Now $E_x$ and $E_y$ denote the approximate initial boundary of the variation of TVA parameters of $\omega_x$ and $\omega_y$ respectively. If $\omega^*$ is denoted as the optimal relaxation factor then

$$E_x = P_x \mid_{\max} = \frac{\omega_y \sim \omega_x}{2(\omega_x + \omega_y)} \tag{15}$$

So that $\omega_x^m = (0.5 + P_x \mid_{\max})(\omega_x + \omega_y) \approx \omega_y$ and

$$E_y = P_y \mid_{\max} = \frac{\omega^* \sim \omega_y}{2(\omega_y - \omega_L)} \text{ or } \frac{\omega^* \sim \omega_y}{\omega_U - \omega_y} \tag{16}$$

so that, $$\omega^* \approx \omega_y^m = \begin{cases} \omega_y + P_y|_{max}(\omega_U - \omega_y), \omega_y > \omega_x \\ \omega_y + P_y|_{max}(\omega_L - \omega_y), \omega_y < \omega_x \end{cases} \quad (17)$$

### 3.4.2. TS selection method

Each slave contains a subpopulation and each subpopulation consists of at least two individuals. After completing adaptation operation, best individual is selected between two consecutive individuals. These individuals are returned back to the master and compete among themselves. After the competition, a best quality champion is selected according to the error value and checking with the standard value. If the error value is equal or less to the standard value, this is the optimized solution. Otherwise this best champion is cloned and fills the archive and continues the process. The corresponding relaxation factors are also rearranged which is required for adaptation operation of next generation.

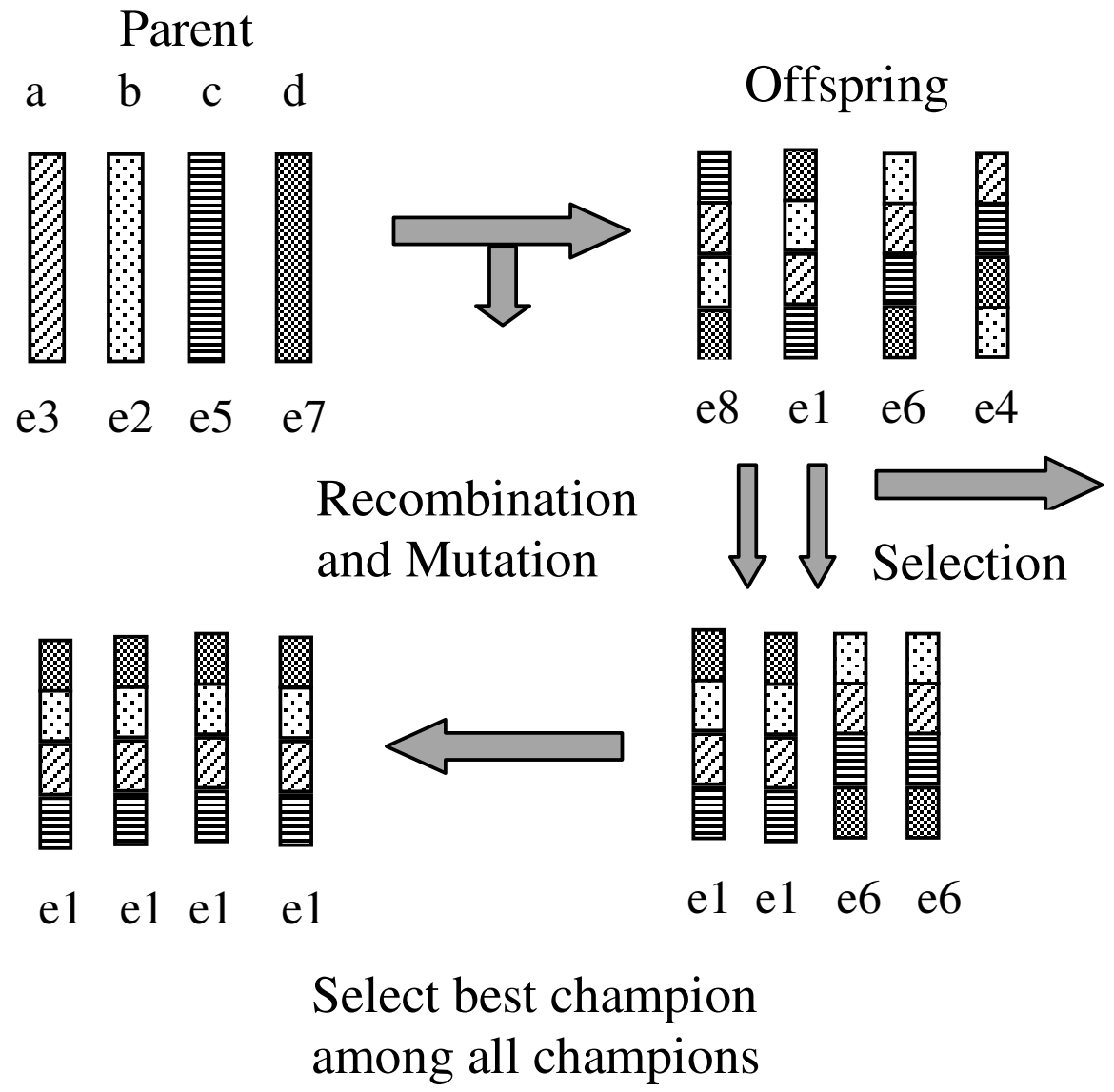

Figure 6. Selection mechanism for TS method

In Figure 6: a, b, c and d patterns represent four parent individuals and the stripped designs represent offspring. Error of each offspring is e8, e1, e6, e4 and corresponding relaxation factors are ω8, ω1, ω6, ω4. There are two subpopulations in two slaves and each contains two individuals. The error rate of the individuals in 1st subpopulation is e1 and e8, and 2nd subpopulation is e6 and e4. Now, best individual is selected between two consecutive offsprings and made a twin of it. The individual with the error rate e1 is chosen from 1st individual and individual with the error rate e6 is chosen from 2nd subpopulation. These individuals are returned back to master and select a best quality champion according to the error rate. This champion is cloned to the equal number of the parent and next generation is started with these cloned individuals. Corresponding relaxation factor is also rearranged.

The time variant adaptive (TVA) parameters are defined as

$$P_x = gauss_x \times P_{max} \times T_\omega \quad (18)$$

$P_x$ is denoted as adaptive (TVA) probability parameter of $\omega_x$, and

$$P_y = gauss_y \times P_{min} \times T_\omega \quad (19)$$

$P_y$ is denoted as adaptive (TVA) probability parameter of $\omega_y$

Where $T_\omega = \lambda \ln\left(1+\frac{1}{t+\lambda}\right), \lambda > 10$

Approximately, $P_{\max}$ and $P_{\min}$ are two exogenous parameters that are assumed as

$$P_{\max} = 0.125 \quad \text{and} \quad P_{\min} = 0.0325$$

Here, $\lambda$ and $\gamma$ are exogenous parameters, used for increased or decreased of rate of change of curvature with respect to number of iterations; $t$ and T denote number of generation and maximum number of generation respectively. Also, $gauss_x(0,0.25)$ is the Gaussian distribution with mean 0 and standard deviation 0.25. Now $P_{\max}$ and $P_{\min}$ denote the approximate initial boundary of the variation of TVA parameters of $\omega_x$ and $\omega_y$ respectively. And if $\omega^*$ is denoted as the optimal relaxation factor then

$$\omega^* \approx \omega_y^m = \begin{cases} \omega_y + P_y \mid_{\max} (\omega_U - \omega_y), \omega_y > \omega_x \\ \omega_y + P_y \mid_{\max} (\omega_L - \omega_y), \omega_y < \omega_x \end{cases} \quad (20)$$

# 4. Computation Time Analysis of Single Processor and Distributed Processors

Timing is the main parameter which is compared for single and distributed processors. For a specific optimization problem, DEC system provides some notation for recombination time, mutation time, fitness evaluation time, adaptation time, and selection time can be denoted as $T_r$, $T_m$, $T_f$, $T_a$ and $T_s$. Total time in single processor is indicated by $T_{\sin gle}$. Time in single processor is represented as follows,

$$T_{Single} = T_r + T_m + T_f + T_a + T_s \quad (21)$$

In case of distributed system, time is calculated in master and slaves separately. Then total time is calculated by combining master and slave time.

## 4.1. Time in master processor

For a particular problem, Recombination operation is performed on initial population and time $T_r$ is calculated. Then the server distributes individuals among all slaves connected to it. Number of subpopulations is $p$ which is distributed in slaves. In this case, Marshalling and transmission time is considered. Marshalling time is the time to construct data stream from object and transmission time is the time to place data stream in channel. Marshalling time of $i^{th}$ individual is $T_{mi}$ and Transmission time with $i^{th}$ individual is $T_{transi}$.

Marshalling and transmission time with $p$ subpopulations $= P[T_{mi} + T_{transi}]$

## 4.2. Time in slave processors

Mutation time, fitness evaluation time, adaptation time with $p$ subpopulations is $T_m$, $T_f$ and $T_a$ respectively. Unmarshalling time $T_{umi}$ is considered here. Unmarshalling time is the time to create object from data stream. After completing adaptation operation, each slave sends mutated offspring to server when BAS selection method is considered but in TS method, slaves send individuals which are selected as best quality for a particular slave. In BAS method, selection operation is performed in master processor but partial selection operation is completed in slave processor in TS method. So this selection time for TS method is calculated in slave processor and sends to master. The processing time of each slave is not equal because of communication delay. In experimental calculation, maximum time is considered for all cases.

### 4.3. Time in Master Processor for selection

The basic difference between BAS and TS selection method is to perform selection operation in two different ways. So selection time for BAS and TS method is calculated in different manner. Selection time with $2N$ subpopulation is $T_s$, $N$ individuals for parent and $N$ individuals for offspring. Maximum slave time is considered for calculating speedup. Consider $m$ is the number of slaves. In BAS selection method, total time for distributed processors is as follows,

$$T_{Distributed(BAS)} = T_r + p\left[T_{mi} + T_{transi} + \max_{1....m}\left(T_m + T_f + T_a\right)\right] + pT_{umi} + T_s \tag{22}$$

Where $T_{distributed}$ represents total distributed time, $T_r$, $T_{mi}$, $T_{transi}$, $T_m$, $T_f$, $T_a$, $T_{umi}$, $T_s$ represents recombination time, marshalling time, transmission time, mutation time, fitness evaluation time, adaptation time, unmarshalling time and selection time.
Speedup for distributed technique using BAS method is:

$$Speedup = T_{Single} / T_{Distributed(BAS)} \tag{23}$$

In TS selection method, selection operation is performed on mutated offspring in each slave where each best individual is selected between two consecutive individuals
and corresponding time is obtained. Then slaves send selected offspring with selection time calculated in slave to master.

Total selection time with $p$ subpopulation = $T_s$

Consider $m$ is the number of slaves. In the case of TS method:

$$T_{Distributed} = Tr + p\left[T_{mi} + T_{transi}\right] + \left[T_f + T_m + T_a + T_s\right] \tag{24}$$

Speedup for TS method:

$$Speedup = T_{Single} / T_{Distributed} \tag{25}$$

Furthermore, percentage of improvement:

%=Speedup/Number of computers (26)

Computation time in distributed processors will be less than single processor. So, speedup will be gradually rising with increasing number of individuals in a population.

## 5. EXPERIMENTAL RESULTS

The environment of experiment consists with 15 homogeneous computers and 100 Mbits/sec Ethernet switch. This system is implemented in Java on a personal computer with Intel Pentium IV and 512MB RAM. In our experiment, individual values are generated randomly. Here, random values are generated using normal distribution with the range from -15 to 15. In order to evaluate the effectiveness and efficiency of the proposed algorithm, various problems are tested. A problem which is tested for different approaches is shown in different graphs. The testing problem is:

Ax = b

Where, $a_{ii} = 20.0$, $a_{ij} \in (0,1)$, $b_i = 10 * i$

and $i = 1,2,3,......,n$, $j = 1,2,3,......,n$

The parameter is $n = 100$ and the problem was solved with an error rate resides in the domain from $9\times10^{-9}$ to $1\times10^{-8}$ for both BAS and TS method. Different experiments were carried out using this problem.

Table I gives the results produced by proposed distributed algorithm and single processor algorithm. Here, different problems are tested with comparing time between single and distributed processors. Table I provides the results using BAS method. Similarly, Table II summarizes experimental results of various problems with TS method. In these two tables, number of computers in a cluster is five as distributed processors. In all cases, optimum solution is achieved. It is possible to solve various benchmark problems using the proposed distributed system.

It is very clear from Table I and Table II that performance of distributed processors is better than single processor. Although BAS and TS both are selection methods that are used in distributed system, BAS method shows better performance than TS method.

Table 1. Experimental Results of Time for Different Problems in Single and Distributed Processors for BAS Method

| Problem Number | Problem | Individual Number | Parameter Number | Time in Single Processor (seconds) | Time in Distributed Processor (seconds) | Error |
|---|---|---|---|---|---|---|
| $p1$ | $a_{ii} = 20.0; a_{ij} \in (0,1);$ $b_i = 10 * i$ | 40 | 170 | $4.02\times10^{-1}$ | $6.88\times10^{-2}$ | $9\times10^{-4}$ |
| $p2$ | $a_{ii} = 20n; a_{ij} = j; b_i$ | 40 | 120 | $3.68\times10^{-1}$ | $6.75\times10^{-2}$ | $9\times10^{-9}$ |
| $p3$ | $a_{ii} = 2i^2; a_{ij} = j; b_i =$ | 40 | 120 | $3.38\times10^{-1}$ | $6.16\times10^{-2}$ | $9\times10^{-9}$ |
| $p4$ | $a_{ii} \in (-100,100);$ $a_{ij} \in (-10,10);$ $b_i \in (-100,100)$ | 20 | 100 | $1.18\times10^{-1}$ | $3.57\times10^{-2}$ | $9\times10^{-9}$ |
| $p5$ | $a_{ii} \in (-70,70);$ $a_{ij} \in (0,7);$ $b_i \in (0,70)$ | 20 | 100 | $1.33\times10^{-1}$ | $3.45\times10^{-2}$ | $9\times10^{-9}$ |
| $p6$ | $a_{ii} = 70;$ $a_{ij} \in (-10,10);$ $b_i \in (-70,70)$ | 40 | 100 | $1.85\times10^{0}$ | $2.13\times10^{-2}$ | $9\times10^{-4}$ |

Table 2. Experimental Results of Time for Different Problems in Single and Distributed Processors for TS Method

| Problem Number | Problem | Individual Number | Parameter Number | Time in Single Processor (seconds) | Time in Distributed Processor (seconds) | Error |
|---|---|---|---|---|---|---|
| $p1$ | $a_{ii} = 20.0; a_{ij} \in (0,$ <br> $b_i = 10 * i$ | 20 | 100 | $1.21\times10^{-1}$ | $9.93\times10^{-2}$ | $1\times10^{-8}$ |
| $p2$ | $a_{ii} = 20n; a_{ij} = j;$ | 20 | 100 | $1.28\times10^{-1}$ | $1.05\times10^{-1}$ | $1\times10^{-8}$ |
| $p3$ | $a_{ii} = 2i^2; a_{ij} = j;$ | 20 | 100 | $1.48\times10^{-1}$ | $1.22\times10^{-1}$ | $1\times10^{-8}$ |
| $p4$ | $a_{ii} \in (-100,100);$ <br> $a_{ij} \in (-10,10);$ <br> $b_i \in (-100,100)$ | 20 | 100 | $1.46\times10^{-1}$ | $1.34\times10^{-1}$ | $1\times10^{-5}$ |
| $p5$ | $a_{ii} \in (-70,70);$ <br> $a_{ij} \in (0,7); b_i \in (0,$ | 20 | 50 | $1.06\times10^{-1}$ | $9.59\times10^{-2}$ | $1\times10^{-4}$ |
| $p6$ | $a_{ii} = 70;$ <br> $a_{ij} \in (-10,10);$ <br> $b_i \in (-70,70)$ | 20 | 10 | $1.12\times10^{-1}$ | $9.94\times10^{-2}$ | $1\times10^{-4}$ |

## 5.1. Speedup comparison between BAS and TS method

To compare speedup, BAS and TS method follows the system standard with 40 individuals for 5 and 10 number of computers in a cluster as well as 30 individuals for 15 number of computers when parameters are 100 for each case.

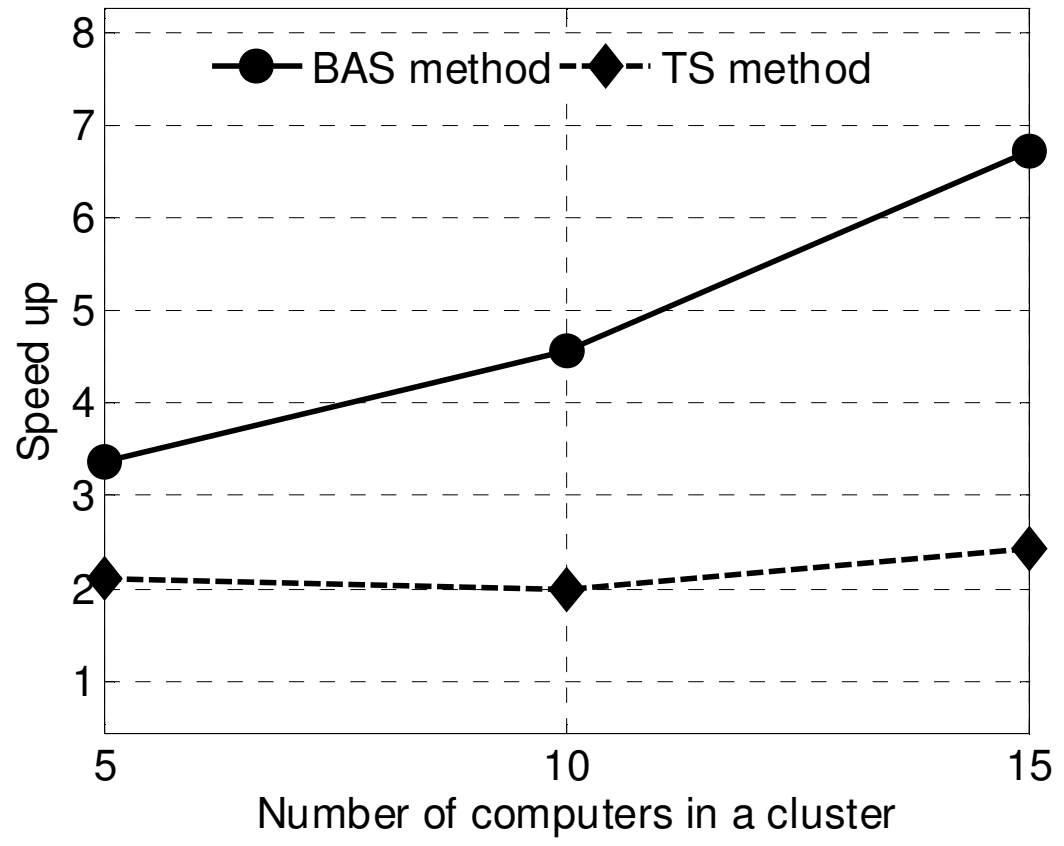

Figure 7. Speed up measurement according to number of computers in a cluster

In Figure 7, speedup is calculated using eqn (23) in BAS method and TS method uses eqn (25). Speedup is 3.36, 4.57, 6.72 in BAS method and 2.109, 1.999, 2.421 in TS method when number of slave computers is 5, 10 and 15. Percentage of improvement is sequentially 67.2 %, 45.7 % and 44.8 % for BAS method as well as 21.09 %, 19.99 % and 24.21 % for TS method which is calculated based on eqn (26). It can be easily visualized that BAS method provides better performance than TS method.

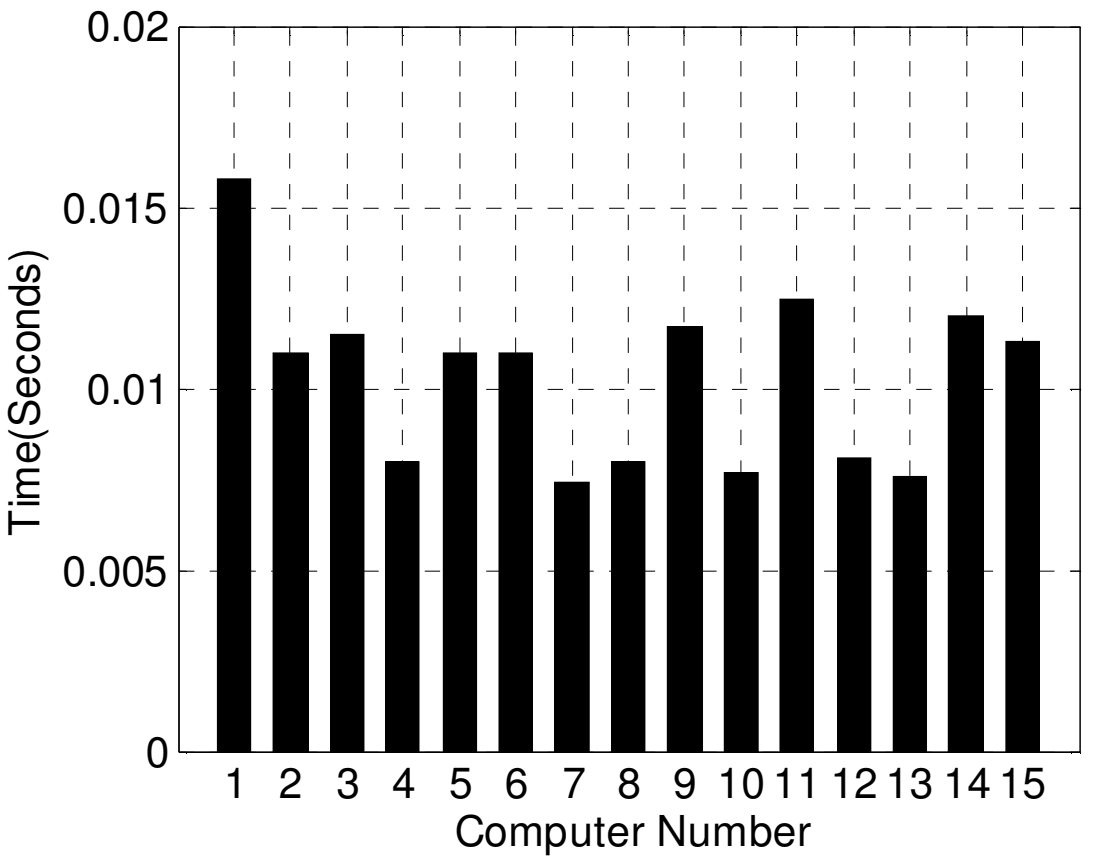

Figure 8. Time (in seconds) measurement according to computer number in a cluster

Figure 8 shows the time required for performing genetic operation in each slave. Here, number of slave computers in a cluster is 15 and individual is 30 and parameters are 100. There is no load balancing system so different slave computers need different time.

Intuitively, workload balancing for a distributed system can be difficult because the working environment in a network is often complex and uncertain. From Figure 8, we can see that computer number 1 needs the highest computation time among all slaves. For calculating total distributed time, this maximum value is considered in both methods.

## 5.2. Comparing time with dimension

Figure 9 and Figure 10 show the time requirement on the basis of parameter number to solve the problems using BAS and TS method and compare the performance in single and distributed processors. The system runs with 5 and 20 slave computers in a cluster and parameters are 50, 60, 70, 80, 90 and 100. There is a fluctuation of time with increasing number of parameters because of random production. All parameters are same for two methods. Distributed system needs less time than single system in both cases but BAS method is better selection mechanism than TS selection method.

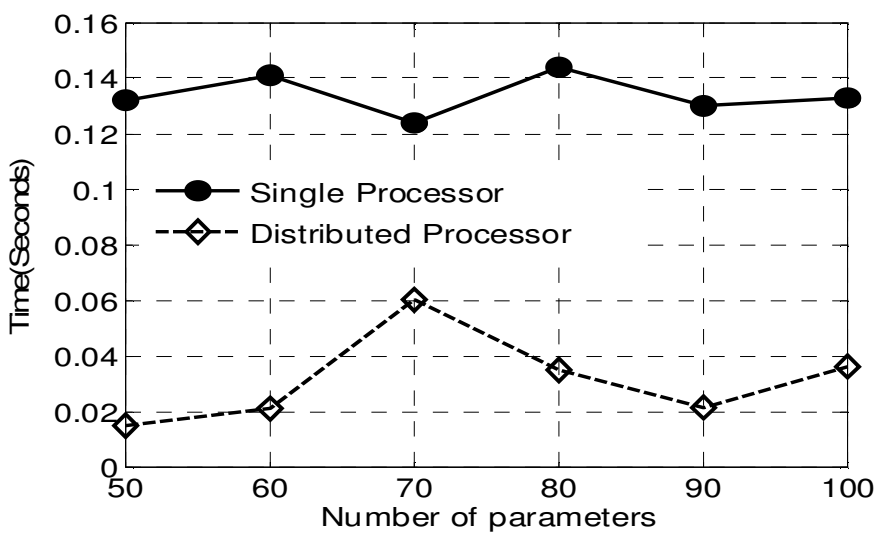

Figure 9. Time measurement according to the number of parameters in BAS method

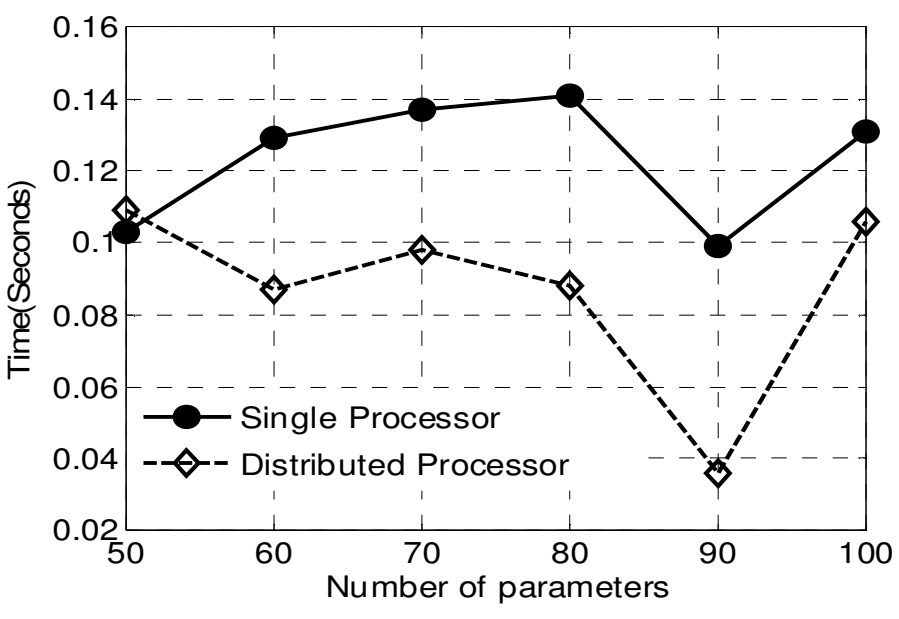

Figure 10. Time measurement according to the number of parameters in TS method

### 5.3. Comparing time with individual

Time is compared between single and distributed processors according to individuals using BAS and TS method; this is shown in Figure 11 and Figure 12. The number of slave computers in a cluster is 5 and number of parameters is 100 in all cases where individual number is varying with the value of 10, 20, 30, 40 and 50. Time is increasing with increasing number of individuals but more time is always needed in single system comparing with distributed system.

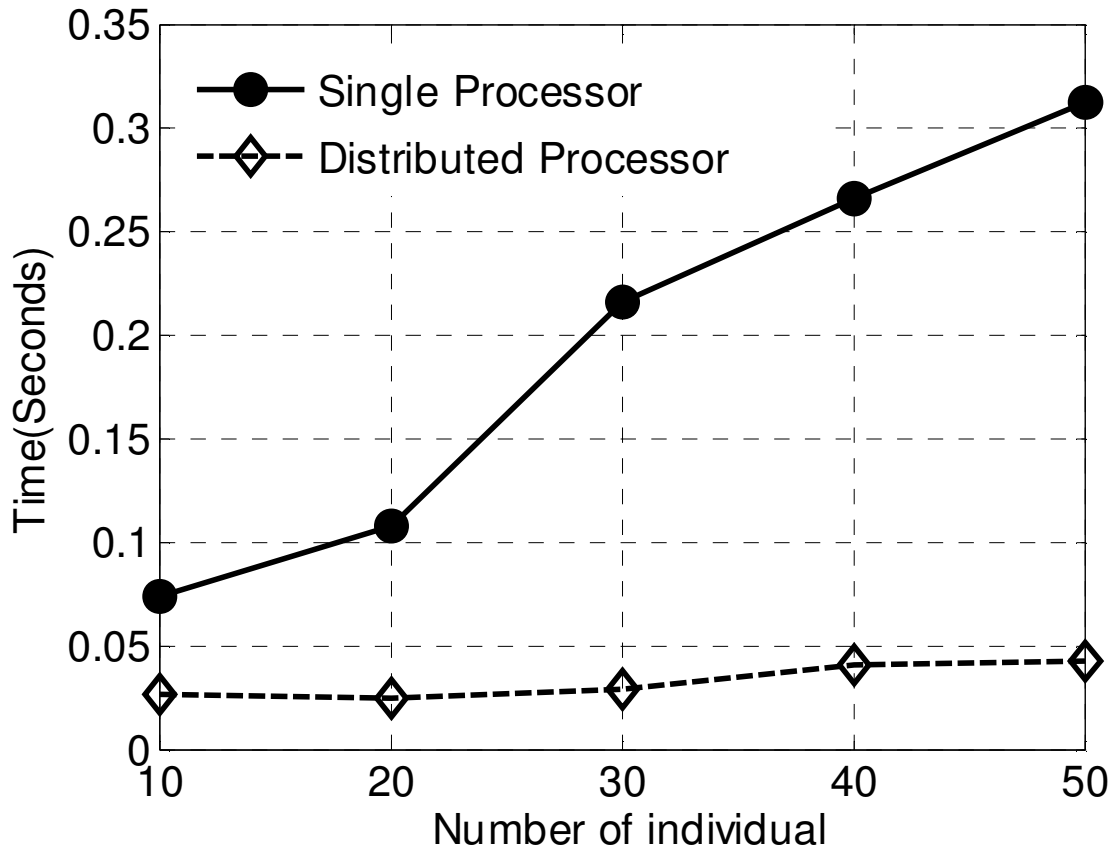

Figure 11. Measurement of Time with number of individuals in BAS method.

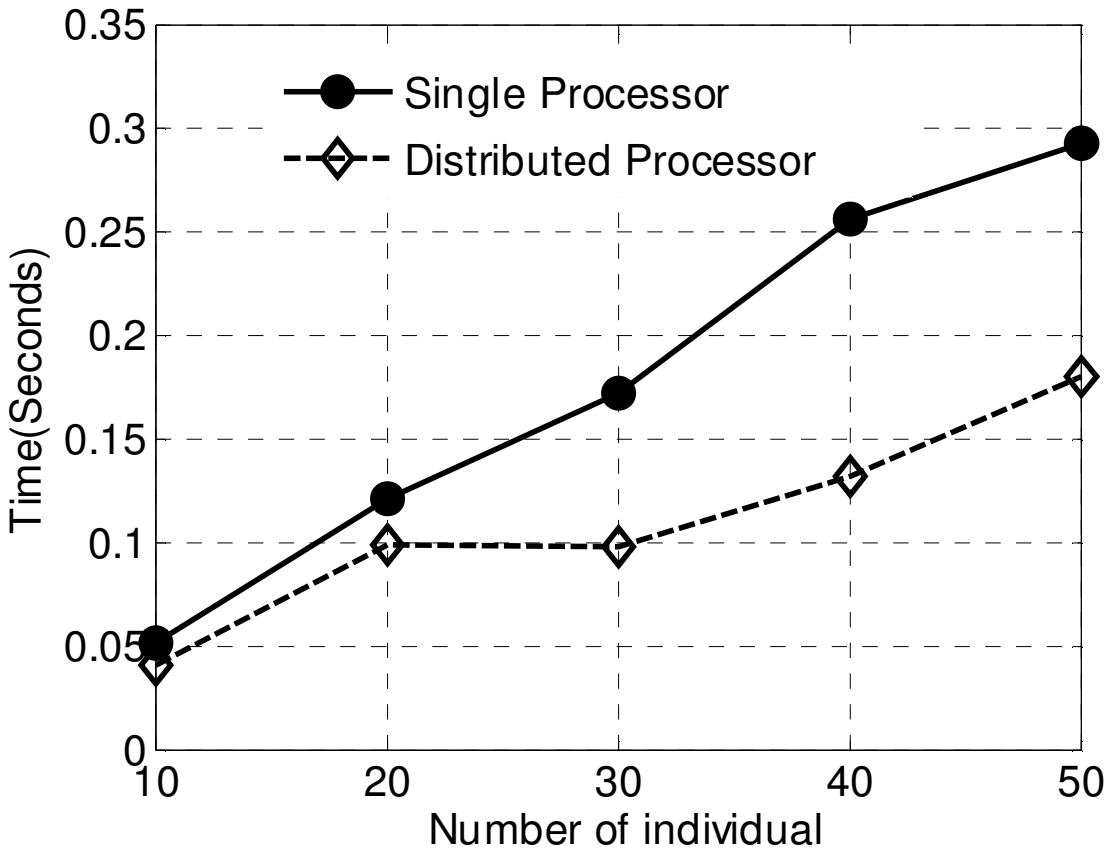

Figure 12. Measurement of Time with number of individuals in TS method

### 5.4. Comparing error with generation

Figure 13 and Figure 14 visualize the error rating with number of generations in BAS and TS method. In both cases, efficiency is compared between single and distributed. These two methods use 5 slave computers in a cluster but other parameters are different. Number of parameters is 100 and number of individuals is 40 for BAS method but TS method uses 90 parameters and 20 individuals for this experiment. From figures it is easily understandable that some cases of distributed system needs less generation than single system to go convergence.

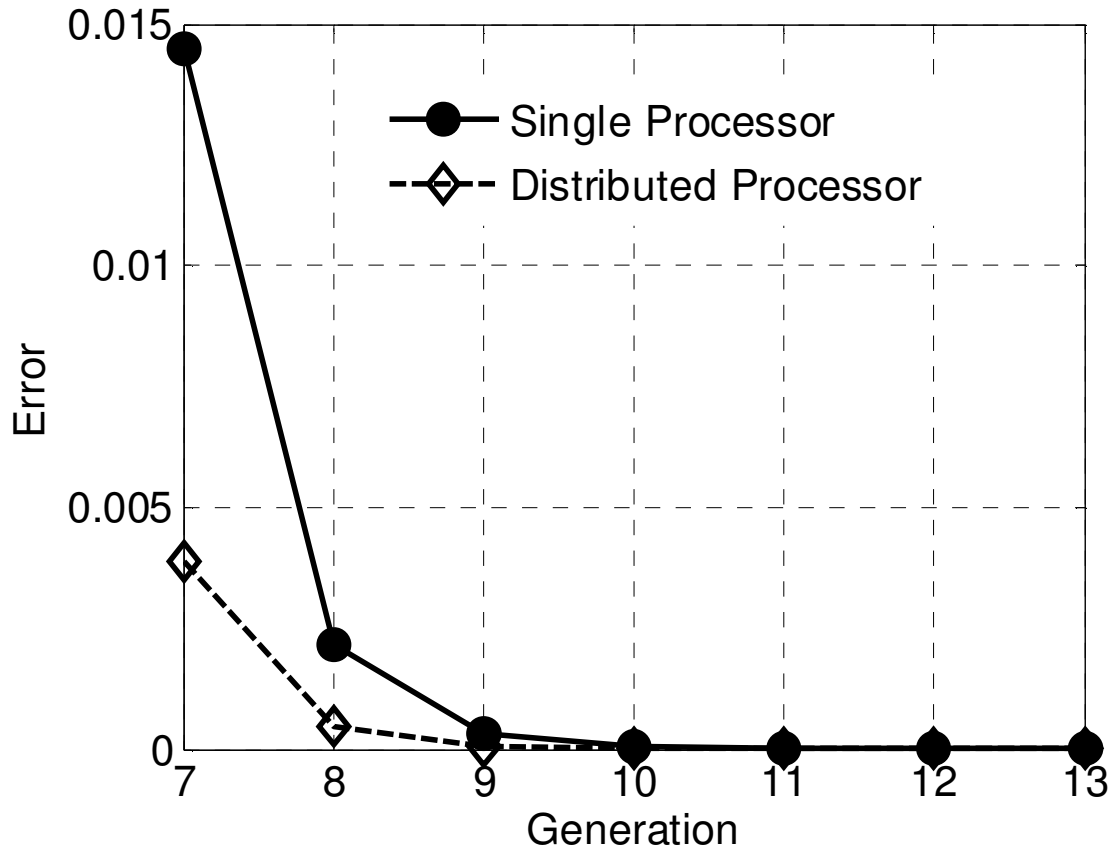

Figure 13. Error measurement according to generation in BAS method

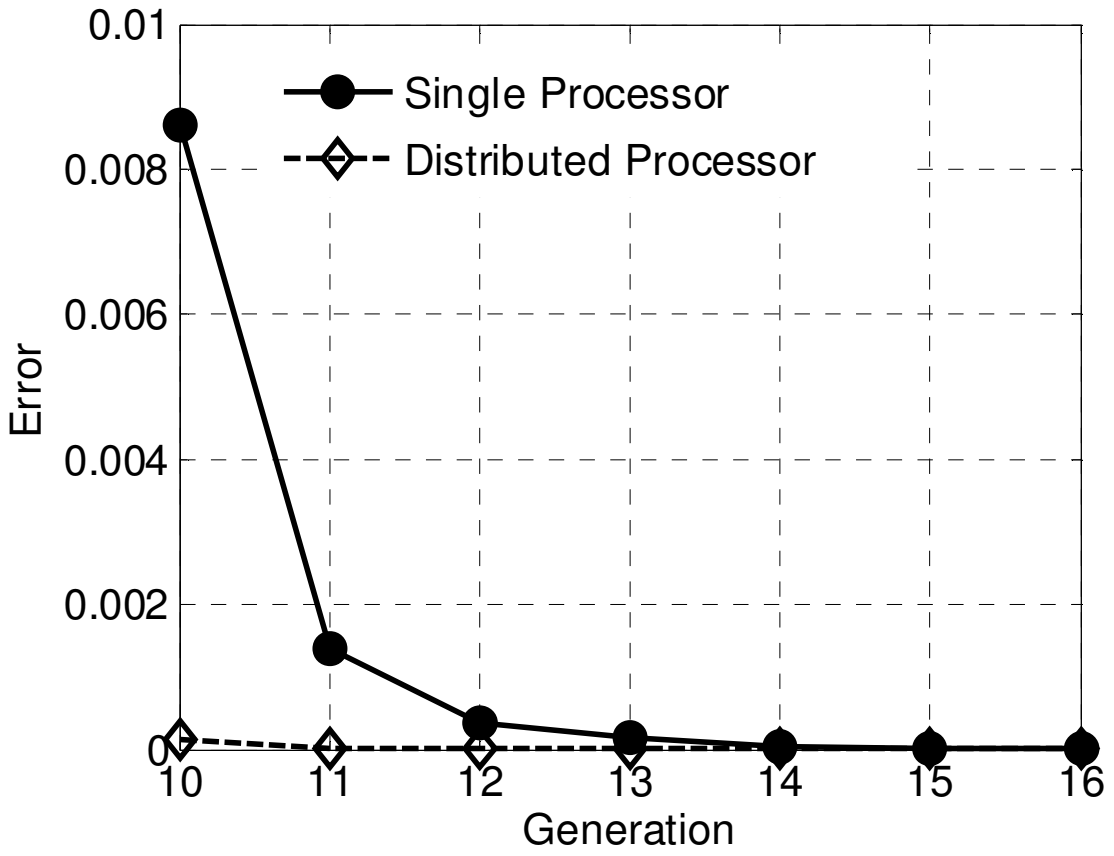

Figure 14. Error measurement according to generation in TS method

In all cases, performance of distributed processor is better than single processor. In the mean time it is also noticeable that BAS selection method shows incredible performance than TS method although this method completes a partial selection in slave processor. A tricky point is that individuals are filtering primarily in each slave, so there may be discarded better quality individual from a slave. On the other hand, BAS method choose best individual from parent and all offspring coming back from all slaves. So achieving optimal result in BAS method is faster than TS method. Apparently, it is not realized because of distributed selection operation in TS method which is not in BAS method. But in some case, TS method provide better performance if the selection order is perfect for each slave.

## 6. CONCLUSIONS

It is easy to solve linear equations with less number of parameters in single processor but high computational cost is required for large number of parameters. This cost can be drastically reduced using distributed system. This paper introduced a new distributed algorithm for solving large number of equations with huge parameters. It also introduced two new selection methods called BAS and TS for selecting offspring. In these methods, best individuals are selected and computation load is distributed in each slave and mutation, fitness evaluation, and adaptation operations are performed on distributed load. As distributed technique is used, computation time is reduced using these selection methods but BAS method provides better performance compared to TS method. For both selection methods, computational time is analyzed and hence speed up is calculated in this new distributed computing system.

**Authors**

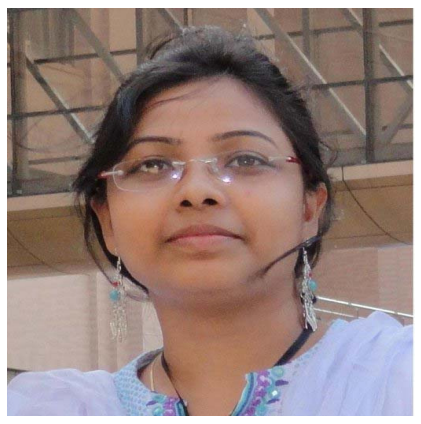

**Moslema Jahan** achieved the B.Sc. Engg. Degree (with honours) in Computer Science and Engineering (CSE) from KhulnaUniversity of Engineering & Technology (KUET), Khulna, Bangladesh. Currently she is serving as a Lecturer in the Department of Computer Science & Engineering in Dhaka University of Engineering & Technology (DUET), Gazipur. Her research interest is on Parallel and Distributed computing, Evolutionary Computation, Networking and Image processing. Moslema Jahan is now Member of the Engineers Institution, Bangladesh (IEB).

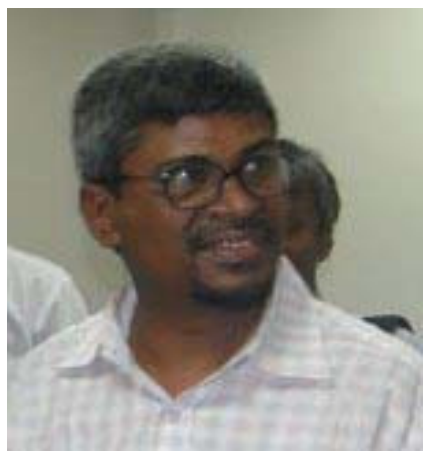

**M. M. A. Hashem** received the Bachelor's degree in Electrical & Electronic Engineering from Khulna University of Engineering & Technology (KUET), Bangladesh in 1988. He acquired his Masters Degree in Computer Science from Asian Institute of Technology (AIT), Bangkok, Thailand in 1993 and PhD degree in Artificial Intelligence Systems from the Saga University, Japan in 1999. He is a Professor in the Department of Computer Science and Engineering, Khulna University of Technology (KUET), Bangladesh. His research interest includes Soft Computing, Intelligent Networking, Wireless Networking, Distributed Evolutionary Computing etc. He has published more than 50 referred articles in international Journals/Conferences. He is a life fellow of IEB and a member of IEEE. He is a coauthor of a book titled "Evolutionary Computations: New Algorithms and their Applications to Evolutionary Robots," Series: Studies in Fuzziness and Soft Computing, Vol. 147, Springer-Verlag, Berlin/New York, ISBN: 3-540-20901-8, (2004). He has served as an *Organizing Chair, IEEE 2008 11$^{th}$ International Conference on Computer and Information Technology (ICCIT 2008) and Workshops, held during 24-27 December, 2008 at KUET.* Currently, he is working as a Technical Support Team Consultant for Bangladesh Research and Education Network (BdREN) in the Higher Education Quality Enhancement Project (HEQEP) of University Grants Commission (UGC) of Bangladesh.

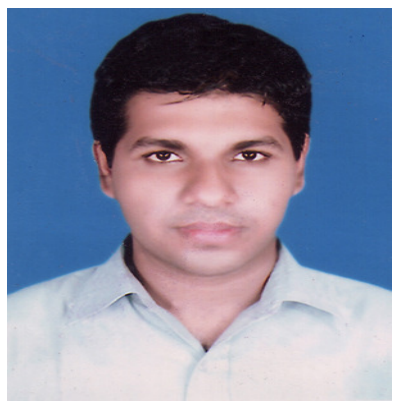

**Gazi Abdullah Shahriar** achieved the B.Sc. Engg. Degree in Computer Science and Engineering (CSE) from Khulna University of Engineering & Technology (KUET), Khulna, Bangladesh. Now he is working at Secure Link Services BD Ltd as software engineer. His current research interest is in Evolutionary computation, Distributed computing and Bioinformatics.